\documentclass[journal]{IEEEtran}

\usepackage{ifthen}
\usepackage{amsmath,bm}
\IEEEoverridecommandlockouts       

\usepackage[utf8]{inputenc}
\usepackage{graphicx}
\usepackage{amsmath}
\usepackage{algorithm}
\usepackage[noend]{algpseudocode}
\usepackage{multirow}
\usepackage{subcaption}
\usepackage{verbatim}
\usepackage{color,soul}

\usepackage{multirow}
\usepackage[dvipsnames]{xcolor}

\newcommand{\mse}{\operatorname{mse}}

\usepackage[switch]{lineno}
 \nolinenumbers

\begin{document}
\title{Optimal Actor-Critic Policy with Optimized Training Datasets}
%
%
\author{Chayan Banerjee$^1$,
Zhiyong Chen$^1$, Nasimul Noman$^2$, and Mohsen Zamani$^{1,3}$
\thanks{$^1$School of Engineering, University of Newcastle, Callaghan, NSW 2308, Australia. 
$^2$School of Information and Physical Sciences,
        University of Newcastle, Callaghan, NSW 2308, Australia. 
        $^3$Department of Medical Physics and Engineering,
Shiraz University of Medical Sciences,  Iran.
        Z. Chen is the corresponding author. Email: {\tt\small  zhiyong.chen@newcastle.edu.au}}%
}
%
%
%
\maketitle              
\begin{abstract}
Actor-critic (AC) algorithms are known for their efficacy and high performance in solving  reinforcement learning problems, but they also suffer from low sampling efficiency. An AC based policy optimization process is iterative and needs to frequently access the agent-environment system to evaluate and update the policy by rolling out the policy, collecting rewards and states (i.e. samples), and learning from them. It ultimately requires a large number of samples to learn an optimal policy. To improve sampling efficiency, we propose a strategy to optimize the training dataset that contains significantly less samples collected from the AC process.  The dataset optimization is made of a best episode only operation, a policy parameter-fitness model, and a genetic algorithm module.  The optimal policy network trained by the optimized training dataset exhibits superior performance  compared to many contemporary AC algorithms in controlling autonomous dynamical systems. Evaluation on standard  benchmarks shows that the method improves sampling efficiency, ensures faster convergence to optima, and is more data-efficient than its counterparts. \end{abstract}
 
\section{Introduction}

Reinforcement learning (RL) has demonstrated significant progress and achieved remarkable performance in diverse domains including robotics \cite{levine2016end,singh2020scalable}, locomotion control \cite{haarnoja2018learning,yang2019data}, strategy games \cite{silver2017mastering,jaderberg2019human}, manufacturing systems \cite{yuan2020general},
and so on. RL algorithms have various choices of learning one or combinations of policies, action-value functions (Q-functions), value functions and/or environment models. In particular, 
actor-critic (AC) algorithms \cite{konda2000actor} are a class of RL algorithms that learn optimal policies. 
In a policy optimization process, an AC algorithm consists of approximate value function estimation,
performance evaluation of the current policy, and policy update. 
Readers can refer to more principles and progresses of optimization methods in machine learning
in a recent survey paper \cite{sun2019survey}.

AC algorithms have been proved to be effective in solving complicated RL problems; see, e.g., \cite{gu2016q,schulman2015high}. 
However, they always suffer from sampling inefficiency because of
the fundamental restriction in using  the on-policy learning approach. 
Roughly speaking, an on-policy approach requires new samples to be collected at
every step of policy update.  
Such a sample collection manner causes substantial increase in cost of experiments
(for real world scenarios) or computation (for simulated environments).
It is worth mentioning that there have been notable works, e.g., \cite{schulman2015trust,wu2017scalable,schulman2017proximal}, for improving the stability and sampling efficiency of AC algorithms in the on-policy framework. 
Trust region policy optimization (TRPO) \cite{schulman2015trust} updates policies by taking the largest possible step, while satisfying a KL-divergence constraint on the closeness of old and new policies. 
A scalable trust region method \cite{wu2017scalable} shows its improvement in sample efficiency.  
Proximal policy optimization (PPO) \cite{schulman2017proximal} replaces the hard KL constraint of TRPO with a penalty on KL divergence
and also proposes an alternative surrogate objective.

Alternatively, off-policy methods have  also been extensively used with improved sampling efficiency.
While an on-policy algorithm learns the value of the policy being carried out by the agent, including the exploration steps, 
an off-policy algorithm learns the value of the optimal policy independently of the agent's actions by executing a separate exploratory policy \cite{sutton2018reinforcement}.  Typical off-policy algorithms include the well known Q learning (QL) \cite{watkins1992q} and related works such as deep Q-network (DQN) \cite{mnih2013playing} and double-DQN  \cite{van2016deep}.
An off-policy version of AC algorithm, Off-PAC, was proposed in \cite{degris2012off}. 
In Off-PAC, the actor executes actions sampled from a fixed behavior policy 
and the critic learns an (off-policy) estimate of the value function  for  the  current  policy. 
The estimate is later used to update the weights of the critic and the policy. 
In \cite{song2015off},  an off-policy integral RL algorithm based on AC networks was developed
for optimal control of unknown systems subject to unknown disturbances 
with the aid of  a disturbances compensation controller.

Off-policy algorithms also employ a technique called experience replay \cite{lin1992self}. The concept of experience replay involves storing the agent's experience in a dataset. Then, mini-batches of samples from the experience dataset are drawn uniformly at random
for a learning process. It logically separates the process of gaining experience and learning and has been proved 
to be effective in increasing sampling efficiency \cite{mnih2015human}. For instance, the concept was used in 
deep deterministic policy gradient (DDPG) \cite{lillicrap2015continuous} that concurrently learns a Q-function and a policy 
and uses the Q-function to update the policy. However, DDPG is sensitive to hyper-parameters and may cause overestimation of the learned Q-function. Then,  a twin delayed DDPG (TD3) was proposed in \cite{fujimoto2018addressing} to address this overestimation issue. 
The concept of experience replay was also used in AC algorithms; see the 
AC with experience replay (ACER)  algorithm introduced in  \cite{wang2016sample}.
It is worth mentioning that simple and convenient implementation of an off-policy adaptive QL method was developed  
in \cite{luo2018adaptive}. In particular, the experience replay technique is employed in the learning process in an AC neural network structure.

The methods  for improving sampling efficiency of AC algorithms are not limited to 
those discussed above.  
For example, the soft actor-critic (SAC) algorithm introduced in \cite{haarnoja2018soft} 
is another effective method that  is based on the concept of entropy regularization. 
In SAC, the policy is trained to maximize the trade-off between expected return and entropy. 
Research showed that sampling efficiency of SAC exceeds that of DDPG and other benchmarks by a substantial margin.
 
Other than the aforementioned on-policy and off-policy approaches, an offline or batch learning \cite{lange2012batch} 
approach has also gained researchers' interest,  where an experience buffer is maintained like off-policy RL but it is not updated actively from online interaction. It uses previously collected agent-environment interaction data to train policies \cite{fu2020d4rl}. Some recent research using this data-driven paradigm for learning policies includes learning of navigation skills in mobile robots \cite{kahn2021badgr}, learning of human preferences in dialogue \cite{jaques2019way}, and learning of robotic manipulation \cite{kalashnikov2018scalable,zeng2018learning}. Readers can refer to a more detailed and holistic coverage of offline learning in a recent survey paper \cite{levine2020offline}.
 
The proposed approach in this paper sits on the intersection of offline and off-policy learning approaches. Similar to offline learning, our approach learns from a static dataset, which is collected by running some prior policy but not continually updated (unlike off-policy algorithms). Furthermore, unlike most offline algorithms where a model-based RL setup is used, our work is completely model free, like off-policy methods.
More specifically, the new approach is along with the research line of improving sampling efficiency of AC algorithms, 
especially using separated process of gaining experience and learning. 
We propose a strategy to optimize the experience dataset before it is
used as a training dataset for learning an optimal policy. As a result, the training dataset requires 
 significantly fewer samples collected from the AC process.  
 Such a dataset optimization process is made of a best episode only operation, a policy parameter-fitness model,
and a genetic algorithm (GA) module.  
The optimal policy network trained by the optimized training dataset exhibits superior 
performance compared with the conventional AC algorithm. 
Evaluation on standard benchmarks shows that the method improves sampling efficiency, ensures faster convergence to optima, 
and is more data-efficient than its counterparts.  
 Since  a GA module is used for optimizing the training dataset collected from an AC process, 
the algorithm in this paper is called a genetic algorithm aided actor-critic (GAAC).

It is worth noting that GA, as a class of evolutionary algorithms,
has been successfully used as an alternative to RL \cite{gangwani2017genetic,such2017deep} 
or as an aid to improve the performance of RL \cite{khadka2018evolution,khadka2019collaborative}. 
For instance,  in \cite{khadka2018evolution} evolution based learning is incorporated with  RL's gradient based optimization in a single framework to maintain a best policy population for evaluation and eventual convergence to an optimal policy. 
A collaborative evolutionary RL (CERL) was proposed in  \cite{khadka2019collaborative} 
which enables collective exploration of policies by policy gradient and neuroevolution modules to evolve an optimal policy network. GA was also used  to optimize hyper-parameters of RL algorithms in \cite{sehgal2019deep} and
evolve neural network weights \cite{sun2019evolving,hameed2019using}.   
A GA based  adaptive momentum estimation (ADAM) algorithm,  
called genetic-evolutionary ADAM (GADAM),
learns better deep neural network models based on a number of unit models over generations  \cite{zhang2018gadam}.  

The remaining sections  of the paper are organized as follows. In Section II, we present the preliminaries and motivation of this paper. In Section III, we explain the proposed GAAC approach with optimized training datasets in details. 
In Section IV, we further discuss the GA module used in the dataset optimization process.  
Section V verifies the effectiveness of our approach in term of its comparison with other existing benchmarks. 
Finally, Section VI concludes the paper and discusses  some related future research avenues. 
 
\section{Preliminaries and Motivation}

The paper is concerned about control of autonomous systems in a
Markovian dynamical model  represented by a
conditional probability density function $p(s_{t+1}|s_t,a_t)$
where  $s_t  \in \mathcal{S}$ and $a_t \in \mathcal{A}$ 
are the current state and control action respectively at time instant $t = 1,2,\cdots$, 
and  $s_{t+1} \in \mathcal{S}$ represents the next state at $t+1$.
Here, $\mathcal{S}$ and $\mathcal{A}$ represent the continuous state and action spaces, respectively.
The objective is to learn a stochastic policy $\pi_{\phi}(a_t|s_t)$ parameterized by $\phi$.
Now, the closed-loop trajectory distribution for the episode $t=1,\cdots, T$ can be represented by 
\begin{align*}
   p_{\phi}(\tau) = &{p_{\phi}(s_1,a_1,s_2,a_2,\cdots,s_T, a_{T},s_{T+1})} \\
  = & p(s_1)\prod_{t=1}^T 
     \pi_{\phi}(a_t|s_t) p(s_{t+1}|s_t,a_t) 
\end{align*}
Denote  $r_t = R(a_t, s_{t+1})$ as the reward generated at time $t$. 
The objective is to find an optimal policy, represented by the parameter
 \begin{align*}
     \phi^* = \text{arg max}_{\phi}\, \underbrace{{\mathbf E}_{\tau \sim p_{\phi}(\tau)}\, \Big[\Sigma_{t=1}^T\,R(a_t,s_{t+1})\Big]}_{J(\phi)},
     \end{align*}
which maximizes the  objective function $J(\phi)$.

\begin{figure}[t]
    \centering
        \begin{subfigure}{\columnwidth}
    \includegraphics[width=1\columnwidth]{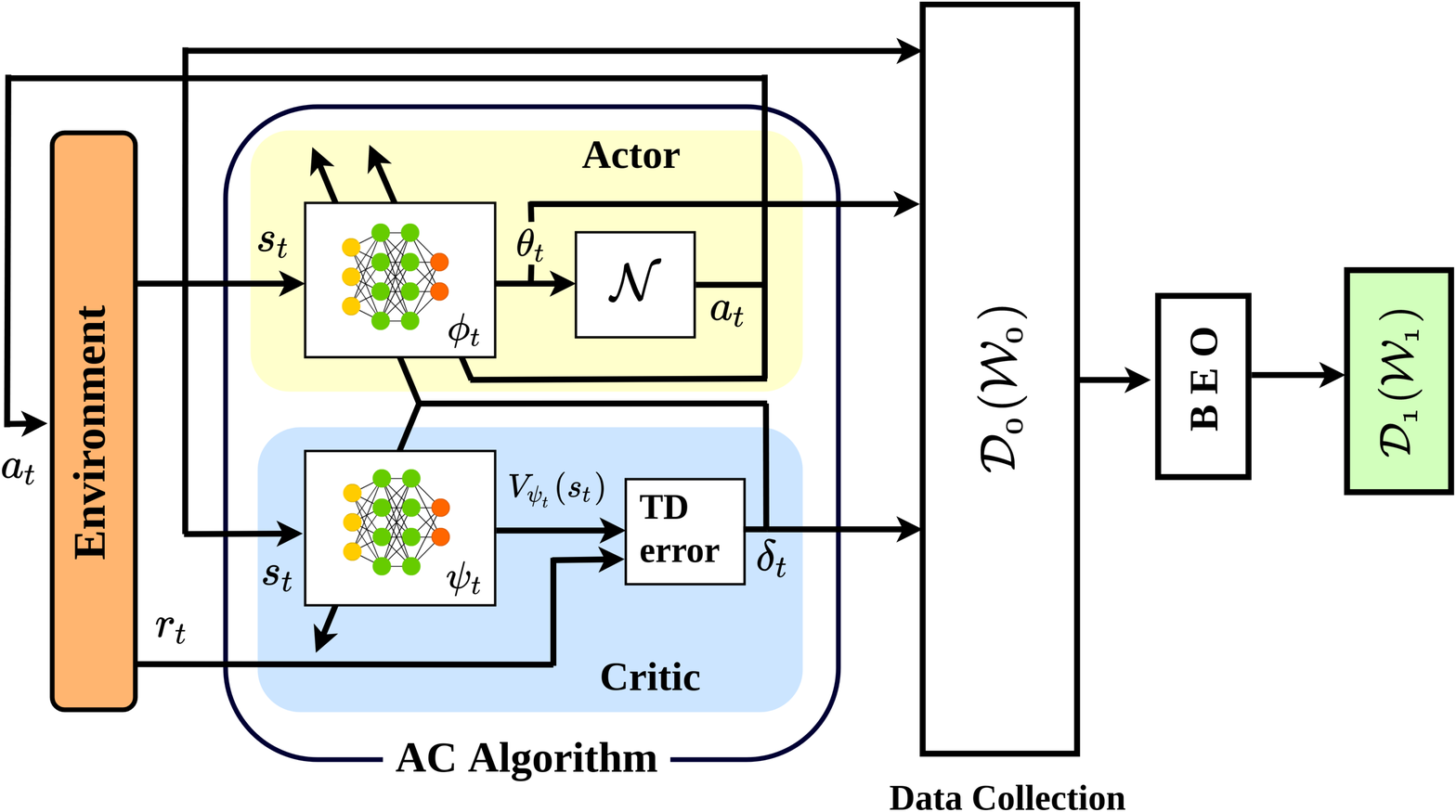} 
        \caption{}
\end{subfigure}
    \begin{subfigure}{\columnwidth}
    \centering
    \includegraphics[width=1\columnwidth]{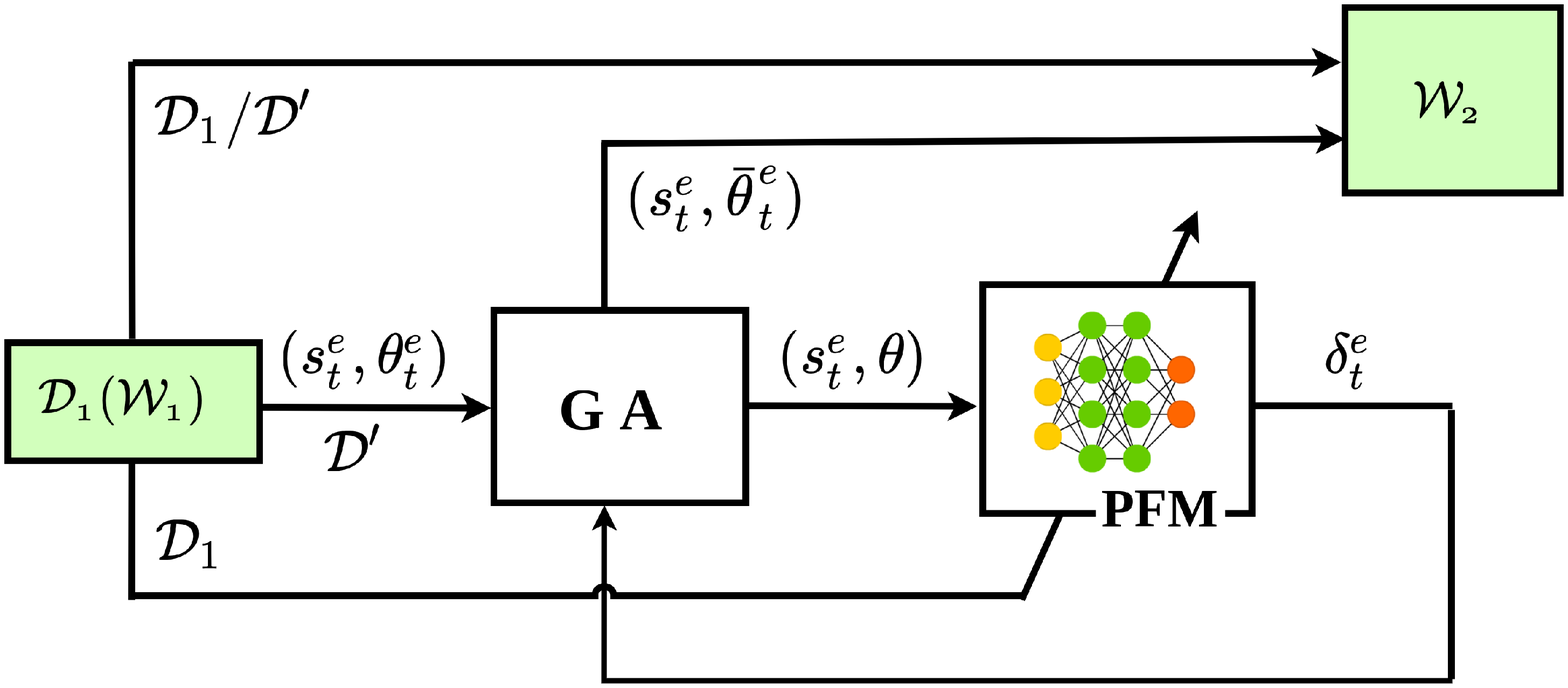} 
    \caption{}
\end{subfigure}
        \begin{subfigure}{\columnwidth}
        \vspace{4mm}
            \hspace{2mm}
    \includegraphics[width=1\columnwidth]{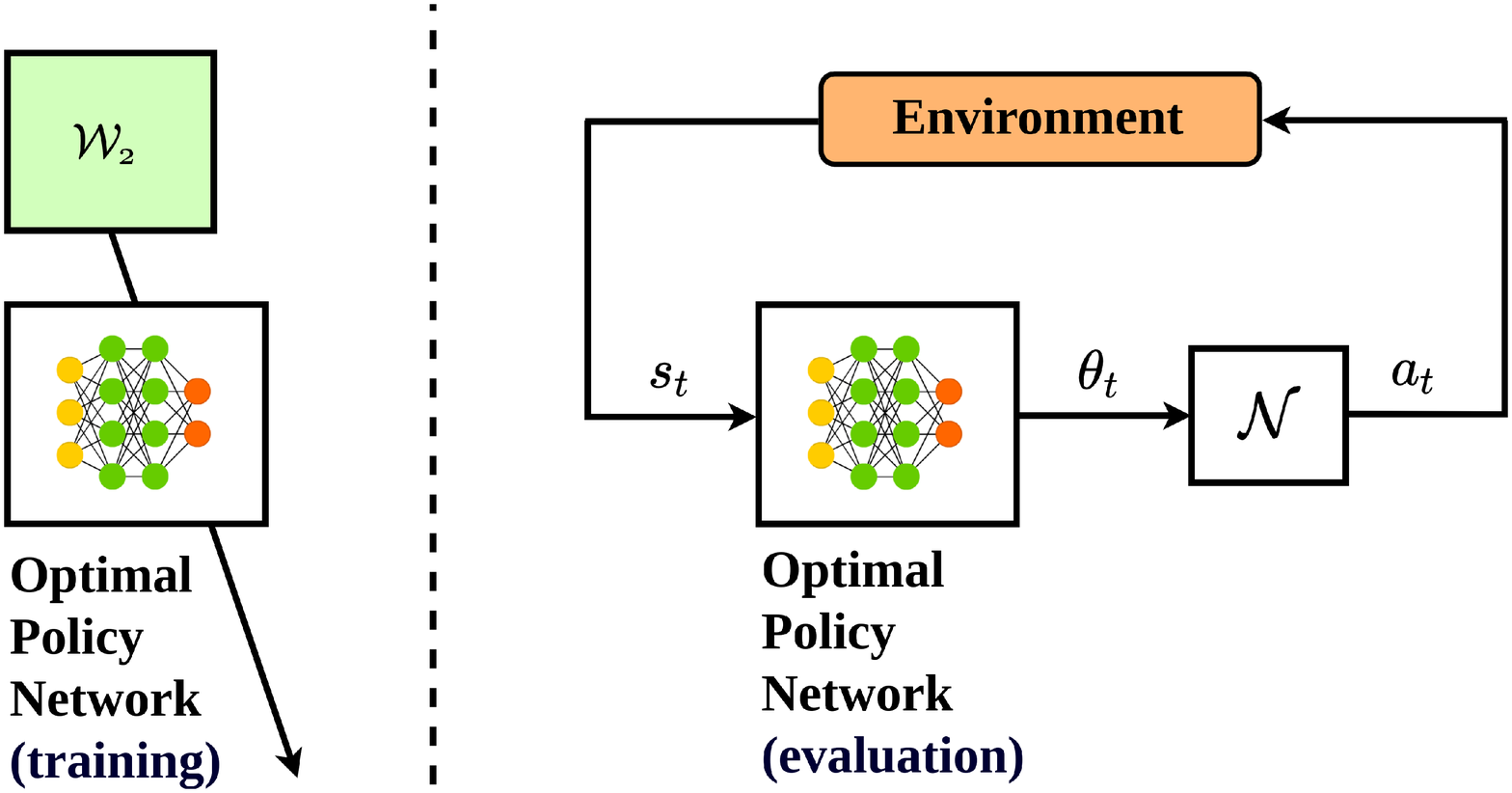}
        \caption{}
\end{subfigure}\caption{Block diagrams of 
    the AC policy with optimized training datasets. (a) Stage 1; (b) Stage 2; (c) Stage 3.}
    \label{fig.diagram}
\end{figure}

We first revisit the conventional AC algorithm that is a class of
model free RL algorithms for achieving the above optimal policy. Later, some improvements will be proposed in this paper. 
The AC Algorithm  is a hybrid of a value based method and a direct policy optimization method \cite{konda2000actor}. A simplified diagram of the AC method is given in Fig.~\ref{fig.diagram}(a). The AC algorithm runs on two function approximators, the actor and the critic, generally modeled using neural networks (NNs).

The critic evaluates the current policy, as reflected by the updated state $s_{t+1}$
and the reward $r_{t}$ received after running the action $a_t$ using the temporal difference (TD) learning.  
Let $\psi_t$ represent the parameters (weights) of the critic NN that generates
the  value function  $V_{\psi_t}(s_{t})$ whose target is the expectation of the cumulative future rewards 
\begin{align} \label{TDtarget}
\sum_{k=0}^\infty \gamma^{k} r_{t+k} = r_t + \gamma \sum_{k=0}^\infty \gamma^{k} r_{t+1+k}
\end{align} given $s_t$. 
Here $\gamma$ is the discount factor which determines the importance of rewards obtained from future states 
compared to those of the current state.
As the target is unknown, the TD learner uses $r_t +\gamma V_{\psi_t}(s_{t+1})$ as the target
of $V_{\psi_t}(s_{t})$, noting the relationship \eqref{TDtarget}.
Therefore, one can define the TD error as
\begin{align}
    \delta_t = r_t +\gamma V_{\psi_t}(s_{t+1}) -V_{\psi_t}(s_{t}).
\end{align}
Then, the simplest update of $\psi_t$ can be
\begin{align}
V_{\psi_{t+1}}(s_{t}) \leftarrow V_{\psi_t}(s_{t}) +\alpha_c \delta_t,
\end{align}
by, e.g.,  using a gradient-based approach, where $\alpha_c$ is the learning rate.

The actor provides a probability distribution over all actions for each state, 
from where the action is sampled and run on the system.
More specifically, let $\phi_t$ represent the parameters (weights) of the actor NN that generates
the policy parameters represented by the vector functions
$\mu_{\phi_t} (s_t), \sigma_{\phi_t} (s_t)$ for the given $s_t$. 
Then, it gives the policy  
that follows the   Gaussian  distribution of mean $\mu_{\phi_t} (s_t)$ and standard deviation
$\sigma_{\phi_t} (s_t)$, i.e., 
$\pi_{\phi_t}(a_t|s_t)=  \mathcal{N}(\mu_{\phi_t} (s_t), \sigma_{\phi_t} (s_t))$.
The action $a_t$ generates the aforementioned TD error $\delta_t$ and hence an 
update approach for $\phi_t$, e.g.,
\begin{align}
\phi_{t+1} \leftarrow \phi_t +\alpha_a  \delta_t \nabla_{\phi_t}\log(\pi_{\phi_{t}}(s_t)).
\end{align} The idea is to minimize the loss that is the negative log likelihood of the Gaussian policy, 
with learning rate $\alpha_a$.

For the convenience of presentation, we denote 
\begin{align*} \theta_t =\left[
\begin{array}{c}
\mu_{\phi_t} (s_t) \\
\sigma_{\phi_t} (s_t) 
\end{array}
\right].\end{align*}
From above, the AC algorithm recursively  updates the policy parameters
using the observations $d_t = \{ s_t, \theta_t, \delta_t \}$, $t=1, \cdots, T$,
in the scenario that the system is fully observed. 
In particular, it is expected that, with a (very) large $T$, 
the optimal policy can be found as $\phi_{T} \rightarrow \phi^*$. 
A more effective way is to keep each episode reasonably small 
according to the real scenario (e.g., an episode is naturally finished 
when a certain task is achieved) and repeat multiple episodes. 
More  specifically, we collect the data $d_t^{e}, \; t=1,\cdots, T_e,\; e=1,\cdots, E$,
in the sequence  of 
\begin{align*}
d_1^1, d_2^1, \cdots, d_{T_1}^1,\cdots,
 d_1^E, d_2^E, \cdots, d_{T_E}^E \end{align*}
where the  superscript $e$ represents the episode index. 
It is noted that the episode length $T_e$ for each episode is not necessarily the same. 
Then, it is expected that, with a (very) large $E$, 
the optimal policy can be found as $\phi^E_{T_E}\rightarrow \phi^*$. 

From above, a successful AC algorithm usually requires a large amount of costly experimental samples. 
It motivates the proposed approach that relies on  considerably lesser experimental samples
that can be optimized and form a training dataset for an optimal policy network. 
The optimization of training datasets is conducted using the multiple ideas listed below.
 They will be elaborated in the following section. 
As a result, the new algorithm improves the conventional AC algorithm through utilizing the
samples more efficiently and achieving faster convergence to the optimal policy.  
The primary contributions and novelties of this paper are listed below. 

 \begin{itemize}
\item[(i)] Best episode only (BEO): The conventional AC algorithm is conducted on real experiments 
for multiple episodes and a raw training dataset is collected. Only the best episodes (selected from repeated rounds)
in terms of  the associated cumulative rewards will be retained to make the final training dataset.  
    
\item[(ii)]   Parameter fitness model (PFM): A PFM NN is created to generate the TD error for a given state and an
unexplored candidate policy parameter. This functionality is used in in dataset optimization.  

\item[(iii)] Dataset optimization: A certain number of policy parameter vectors in the training data
are updated through comparison with other candidates in their neighborhoods in terms of the TD errors
evaluated by the PFM. The selection of candidates typically follows a GA module.

\item[(iv)] Separate policy networks: One policy network is used for running the AC algorithm during data collection and 
the other as the optimal policy network, trained using the optimized dataset.

\end{itemize}

\section{AC Policy with Optimized Training Datasets}

\label{sec:main}

The new AC policy with optimized training datasets is a three-stage process. 
In each stage, we will explicitly explain how the raw training datasets are collected from experiments
and how they are optimized, represented by the sequence of 
\begin{align*}
\mathcal{W}_o \rightarrow \mathcal{W}_1 \rightarrow\mathcal{W}_2
\end{align*}
as elaborated below.
The schematic diagram of the three stages is illustrated in Fig.~\ref{fig.diagram}. 

\subsection{Stage 1: Data collection and selection of best episode}

The first stage starts with collecting data from running the convention AC algorithm for totally $E$ episodes
that are grouped in $M$ rounds of $N$ episodes per round, i.e., $E=MN$.
\begin{figure}[h]
    \centering
    \includegraphics[width=1\columnwidth]{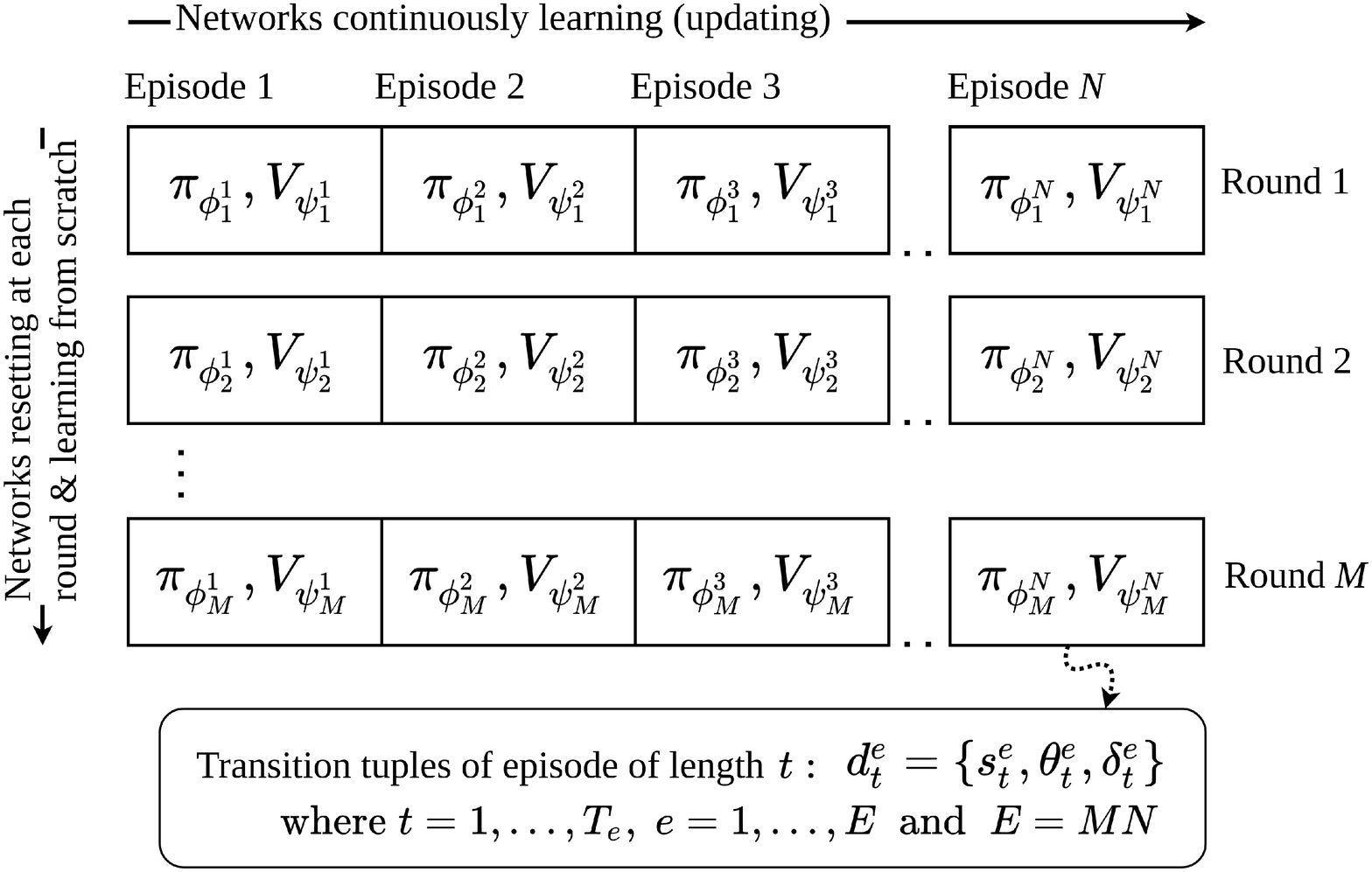}
 \caption{Illustration of the data collection process. }
 \label{fig.stage1_illustrated}
\end{figure}
The data collection stage's setup of $M$ repeated rounds of $N$ episodes is illustrated in Fig.~\ref{fig.stage1_illustrated}.
The raw dataset collected from the experiments is denoted as
\begin{align}
  \mathcal{D}_o = \{d_t^{e} \; | \; t=1,\cdots, T_e,\; e=1,\cdots, E \}.
 \end{align}

Correspondingly, the following input-output pairs 
  \begin{align}
  \mathcal{W}_o = \{ \{ s^e_t, \theta^e_t\} \; | \; t=1,\cdots, T_e,\; e=1,\cdots, E \}.
 \end{align}
 are for the actor NN. In other words, the training dataset for the actor NN is   $\mathcal{W}_o$.
 
Next, for each episode, we define the total reward as 
$r^e =\sum_{t=1}^{T_e} r^e_t$, based on which we can select the best episode 
of each round, that is, 
\begin{align*}
e_m  =  {\arg\max}_{(m-1)N+1 \leq e \leq mN  } \{ r^e \}  ,\;m=1,\cdots, M.
\end{align*}
As a result,  $\mathcal E =\{e_1, \cdots, e_M\}$ is called the set of best episodes. 
It is worth mentioning that, for each round,  both the actor and critic networks are reinitialized
and they start learning from scratch and the learning continues for $N$ episodes.

Then, the dataset from the best episodes, i.e., 
\begin{align}
    \mathcal{D}_1 = \{d_t^{e} \; | \; t=1,\cdots, T_e,\; e \in  \mathcal E \}
\end{align}
 will be used to train the so-called parameter-fitness model in stage 2.
 Similarly, we define 
   \begin{align}
  \mathcal{W}_1 = \{ \{ s^e_t, \theta^e_t\} \; | \; t=1,\cdots, T_e,\; e \in  \mathcal E \}.
 \end{align}
We refer the strategy of selecting the best rewarding episode per round as the BEO approach.
There are two factors which play  crucial role in the efficient learning of a policy from past (collected) experience, i.e., sample balance and data diversity. 
 In a typical training dataset (or replay buffer) the quantity of samples with poor rewards easily outnumbers the quantity of samples with high rewards, leading to sample imbalance, which results in slow policy learning and decreased sample efficiency. Again, training a  policy while considering only the best experience or high reward samples, does not effectively leverage from the policy's exploration behavior. As a result, such a training dataset may suffer from poor data diversity and cause the policy to over-fit and perform
myopically.  

The proposed BEO strategy counters  with these two related issues. In particular, through the BEO approach we retain one ``best in the round'' episode, over multiple $(M = 10)$, short ($N\leq 6$), and mutually uncorrelated rounds of AC policy learning. 
It is worth noting  that there is no fixed threshold for ``best'' episode selection, rather the ``best'' episode selection is from the perspective of each uncorrelated short rounds. This facilitates the selection of even relatively poor rewarding episodes for being the best in a certain round. Therefore, ultimately the policy learns from a training dataset, consisting of data from a selection of episodes with reduced sample imbalance. This enables the algorithm to learn high rewarding policies faster.  Additionally,  these short rounds are mutually uncorrelated and are fresh instances of AC policy learning (networks reset at the onset of each round), which improves exploration behavior leading to enhanced data diversity in the collected dataset $\mathcal{D}_1(\mathcal{W}_1)$.

\subsection{Stage 2: Parameter fitness model and dataset optimization}

The dataset $\mathcal{D}_1$ collected by the AC+BEO method consists of the tuples
$d^e_t = \{ s^e_t, \theta^e_t, \delta^e_t \}$.  In this stage, we first train an NN, called a
parameter-fitness model (PFM), using the training set $\mathcal{D}_1$. 
In particular, the trained NN is represented 
by the function $\rho$ that satisfies 
\begin{align*}
\delta^e_t =  \rho (s^e_t, \theta^e_t), \;\forall d^e_t \in \mathcal{D}_1.
\end{align*}
  The PFM is designed using a multilayered perceptron   and trained to predict the TD error for a given state and  an unexplored candidate policy parameter. This functionality is important for dataset optimization.
We use all data collected in the $\mathcal{D}_1$ as the training data for the PFM network. 
In a supervised learning paradigm, for a sample $d_t^e$ in  $\mathcal{D}_1$, 
the tuple $(s_t^e, \theta_t^e)$ is used as the training input to the PFM network and 
the corresponding $\delta_t^e$ as the target output.
The model is thus learnt by minimizing a mean square error (MSE) loss denoted as $\mse(\hat{\delta_t^e}, \delta_t^e)$ where $\hat{\delta_t^e}$ is the predicted value of the NN under training. The PFM  is further improved by running repeated cross validation tests.

Next, we randomly pick a  subset $\mathcal{D}' \subset \mathcal{D}_1$ of typically $\eta (\%)$ population and optimize
every tuple $d^e_t$ in $\mathcal{D}'$ as follows. 
Define a neighborhood of $\theta^e_t$ as  ${\cal B}(\theta^e_t)$
and find the optimal $\theta$ within this neighborhood in the sense of
\begin{align} \label{optiset}
 \bar \theta^e_t  =  {\arg\max}_{\theta \in {\cal B}(\theta^e_t)} \rho (s^e_t, \theta) , \;\forall d^e_t \in \mathcal{D}'.
\end{align}
Here, $\eta$ is  a  hyperparameter  and  the appropriate  value is  determined  through hyperparameter search,
with more analysis  in Section~\ref{swimerOptbeta}.

Specifically, optimization of \eqref{optiset} is pursued by a GA  module
which is explained in details in the next section. 
This optimization step is crucial since the AC+ BEO based exploratory data in $\mathcal{D}_1(\mathcal{W}_1)$ is quantitatively small and collected from multiple uncorrelated instances (or rounds) of AC policies in their very early  learning process. And so if this data is directly used for training, then it may result in poor performance of the final policy;
see the ablation studies and discussion in Sections \ref{MCC_GAAC artifacts} and \ref{Swim_GAAC_artifacts}. We thus use GA and a surrogate PFM to optimize and update $\eta$ of $\mathcal{W}_1$ before training the optimal policy network. 
For a given state $s^e_t$, the corresponding policy parameter is optimized as $\bar \theta^e_t$
that gives a larger TD error, calculated by the PFM, due to
$\rho (s^e_t,\bar \theta^e_t)>\rho (s^e_t, \theta^e_t) =\delta^e_t$.

For complement of notation, we define
\begin{align}
\bar \theta^e_t  =  \theta^e_t   , \;\forall d^e_t \in \mathcal{D}_1 \backslash \mathcal{D}'.
\end{align}
That is, the policy parameters in the subset $\mathcal{D}_1 \backslash \mathcal{D}'$ are untouched. 
Now, it is ready to have the optimized dataset 
 \begin{align}
 {\cal W}_2 = \{ \{s^e_t, \bar\theta^e_t \}  \; | \;  t=1,\cdots, T_e,\; e \in  \mathcal E \} 
 \end{align}
 that will be used in Stage 3.

\subsection{Stage 3:   Optimal policy training} 

\begin{figure}[h]
    \centering
    \includegraphics[width=1\columnwidth] {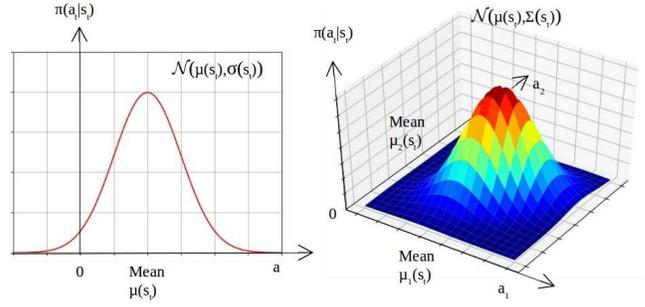}
 \caption{Gaussian policies, with 1D and 2D continuous action space. In 2D case $\mu$ is a vector i.e. $\mu$ = [$\mu_1$,$\mu_2$] and $\Sigma$ is a covariance matrix.}
 \label{fig.gaussian}
\end{figure}

In the final stage,  an optimal policy NN is trained using the dataset  ${\cal W}_2$.
Since a continuous state stochastic policy is concerned, the actions that are sampled from the policy come from a probability distribution given the state. Similar to the mixture density network  concept as introduced in \cite{bishop1994mixture}, the optimal policy NN predicts a mean $\mu$ and a standard deviation value $\sigma$, which define a Gaussian distribution. 
In particular, it is of the same structure as the actor NN whose trained parameters are represented by 
$\phi^*$  satisfying  
\begin{align*} \bar\theta^e_t=\left[
\begin{array}{c}
\mu_{\phi^*} (s_t^e) \\
\sigma_{\phi^*} (s_t^e) 
\end{array}
\right],\; \forall \{s^e_t, \bar\theta^e_t \} \in {\cal W}_2.\end{align*}
The final optimal policy is 
$\pi_{\phi^*}(a_t|s_t)=  \mathcal{N}(\mu_{\phi^*} (s_t), \sigma_{\phi^*} (s_t))$.
When implemented, for any given state, the trained optimal policy NN
is able to give a set of policy parameters that implies an action distribution
and an action sample. A more specific expression of the Gaussian distribution is as follows
\begin{align*}
    \pi_{\phi^*}(a_t|s_t) = \frac{1}{\sqrt{2\pi \sigma_{\phi^*} ^2(s_t)} }\exp\left[-\frac{(a_t-\mu_{\phi^*} (s_t))^2}{2\sigma_{\phi^*} ^2(s_t)}\right].
\end{align*}
which is graphically illustrated in Fig.~\ref{fig.gaussian}.

In other words, the optimal policy NN learns a function   $f: S\rightarrow \bar{\Theta},\; \text{s.t.}\; s_t^e \in S, \, \bar{\theta}_t^e\in \bar{\Theta}.$ For a sample tuple ($s_t^e, \bar{\theta_t^e}$) from the updated dataset $\mathcal{W}_2$, we use $s_t^e$ as the training input to the network and $\bar{\theta}_t^e$ as the target output.  The network is trained by minimizing an MSE loss  given as $\mse(\hat{\theta}_t^e, \bar{\theta_t^e} )$ where $\hat{\theta}_t^e$ is the predicted NN parameter during training.

 \section{Discussion about the GA Module}

\label{sec:GA}
The GA module used in stage 2, aiming at the optimization of \eqref{optiset}, 
is elaborated in this section. The GA carries on through its generic operations like selection of best candidates or parents for mating, based on the TD error predicted by the PFM.
Given a state $s_t^e$, the GA process starts its evolution for the optimal policy parameters from a  batch of candidate optimal parameters, called the initial population, generated around  $\theta^e_t$ collected in the raw dataset, 
followed by crossover and mutation.  Repeating the process over certain iterations (or generations), GA is expected to 
 deliver the optimal policy parameter $ \bar \theta^e_t$ (or pseudo-optimal since the GA is not guaranteed to generate the optimal parameter), as a solution to \eqref{optiset}.

The GA module adopted in this paper is based on GADAM \cite{zhang2018gadam}  with modifications where needed. 
GADAM  was originally proposed as a method for fast optimization of deep NN models. It considers multiple models that have been optimized by an ADAM optimizer and then uses a GA routine to evolve a model with the best possible model parameters.  
Some specific discussion about the GA module is given below.

\subsection{Optimization in the sense of  TD error}

The PFM function $\rho$ is trained to generate the TD error for a given state and  a policy parameter.
So, the optimization of \eqref{optiset} aims to maximize the TD error by selecting the optimal policy parameter.
Intuitively, a TD error quantifies  how much better it is to take a specific action, compared to the average action at the given state. 
For the critic NN, the target is to make the TD error to zero for a good reward evaluation. 
However, for the actor side, a large TD error for a specific action means it brings a higher reward. 
Therefore, in the optimization of \eqref{optiset}, the TD error is considered as a fitness value and a better fitness value means better performance by the policy parameterized by the parameter vector $\bar\theta^e_t$. 

\subsection{Initial population}

Given a state $s^e_t$, the GA process starts its evolution for the optimal policy parameters from a  batch of candidate optimal parameters $\mathcal{G}^{(0)} = \{\theta_1^{(0)}, \cdots,\theta_{J}^{(0)}\}$ (called the initial population) generated in the neighborhood  ${\cal B}(\theta^e_t)$, i.e.,
$\mathcal{G}^{0} \subset {\cal B}(\theta^e_t)$. The initial population size of $\mathcal{G}^{0}$  is denoted as $J$.
 The neighborhood of a parameter $\theta_t^e$ is defined as
$\mathcal{B}(\theta_t^e) = [\theta_{\min}, \theta_{\max}]$ where  
 $\theta_{\min}$ and $\theta_{\max}$ are the minimum and maximum values of all the $\theta_t^e$ collected in $\mathcal{D}_1(\mathcal{W}_1)$, respectively.
 
 To encourage exploration, the initial population $\mathcal{G}^{0}$ is generated as a combination of two separately obtained but equal sized sub populations, i.e.,  
$\mathcal{G}^{0} =  \mathcal{G}_l^{0} \cup \mathcal{G}_m^{0}$.
In particular,  the subset $\mathcal{G}_l^{0} = \{\theta_1^{(0)}, \cdots,\theta_{J/2}^{(0)}\}$ contains the parameters $\theta_j^{(0)}$   randomly selected from 
a set of discrete values $\{  \theta_{\min},
\theta_{\min}+\epsilon, \theta_{\min}+2 \epsilon, \cdots, \theta_{\max}\}$ where 
$\epsilon = (\theta_{\max} - \theta_{\min} )/\ell$ for some integer $\ell >1$
is used to characterize the resolution of the selected parameters. The value of $\epsilon$
is close to $0.05$ in the  experiments of this paper.
The other subset 
$\mathcal{G}_l^{0} = \{\theta_{J/2+1}^{(0)}, \cdots,\theta_{J}^{(0)}\}$
contains the parameters sampled from the distribution  ${\mathcal{N}(\theta_t^e,0.1)}$ and truncated to fit into $\mathcal{B}(\theta_t^e)$.

The optimization process uses the PFM to generate the fitness values for this batch of candidate population, i.e., 
$\rho (s^e_t, \theta), \;\forall \theta \in  \mathcal{G}^{0}$, so that the optimal policy parameter from the initial population can be identified.

\subsection{Selection of parents}

GA learns/evolves the optimal parameters from the initial parameter population. We represent the generation as $\mathcal{G}^{(u)}$ with the candidate parameter population given as $\mathcal{G}^{(u)} = \{\theta_1^{(u)},\cdots,\theta_{J}^{(u)}\}$, where $u\geq 0$ is the number of generations. The corresponding fitness value predicted by the PFM
is given by $\delta_i^{(u)}=    \rho (s^e_t,\theta_i^{(u)})$, $i=1,\cdots, J$.
Then, the selection probability of the unit parameter vector $\theta_i^{(u)}$ as a candidate parent is defined using the following Softmax equation
\begin{align}
 p_i =\frac{\exp{(\delta_i^{(u)})}}{\Sigma_{j=1}^{J} \exp{(\delta_{j}^{(u)})}} . \label{eq:12}
\end{align}
Let $L$ be the number of  parent-pairs and pick 
$\mathcal{G}^{(u)}_1 =\{ \theta_{i_1}^{(u)},\cdots, \theta_{i_L}^{(u)} \}$
as a subset of  $\mathcal{G}^{(u)}$ and $\mathcal{G}^{(u)}_2 = \mathcal{G}^{(u)} \backslash \mathcal{G}^{(u)}_1$, 
such that $p_i \geq p_j$ for all $\theta_{i}^{(u)} \in \mathcal{G}^{(u)}_1$ and $\theta_{j}^{(u)} \in \mathcal{G}^{(u)}_2$.
In other words, $\mathcal{G}^{(u)}_1$ consists of the $L$ parameter vectors of the highest selection
probability. Next, 
we randomly re-order the sequence $i_1, \cdots, i_L$ as $j_1, \cdots, j_L$ such that $i_q \neq j_q$, $q=1,\cdots, L$.
Then, the set of parent pairs is defined as $\mathcal{P}= \{(\theta_{i_1}^{(u)},\theta_{j_1}^{(u)}), \cdots,(\theta_{i_L}^{(u)}, \theta_{j_L}^{(u)})\}$.

\subsection{Crossover}
 
 In GA, the off-spring inherit genes (vector elements) from parents in the crossover process, which propagates the better traits of parents to their children. The child parameter vector thus generated by the crossover process can be represented as $\hat{\theta}_{q}^{(u)}$ from the pair $(\theta_{i_{q}}^{(u)}, \theta_{j_{q}}^{(u)})\in \mathcal{P}$ for $q=1,\cdots, L$.
Let $[h]$ be the $h$-th element of a vector. More specifically, the crossover process is represented by 
\begin{align*}
    \hat{\theta}_{q}^{(u)}[h] = {\rm bool}({\rm rand} <= p_{i_{q},j_{q}}) {\theta}_{i_{q}}^{(u)}[h] \\ + {\rm bool}({\rm rand} > p_{i_{q},j_{q}}) {\theta}_{j_{q}}^{(u)}[h]
\end{align*}
Here, ${\rm bool}$ represents a binary function that returns $1$ if the condition is satisfied and $0$ otherwise, ${\rm rand}$ denotes a random number in $[0,1]$, and $p_{i_{q},j_{q}} = p_{i_q} / (p_{i_q} +p_{j_q})$ is the relative probability.
Obviously, the larger $p_{i_q}$ relative to $p_{j_q}$, the higher chance that the element parameter from the 
parent  ${\theta}_{i_{q}}^{(u)}$ is inducted into the child $\hat{\theta}_{q}^{(u)}$.
After the crossover process, the children population is generated as   $\hat{\mathcal{G}}^{(u)} = \{\hat{\theta}_1^{(u)}, \cdots, \hat{\theta}_L^{(u)}\} $.

\subsection{Mutation}

To avoid trapping in local optima, GA uses a mutation operation. In this process, we introduce randomness into the child parameter vector to encourage exploration. For each child parameter vector $\hat{\theta}_{q}^{(u)}$,  $q=1,\cdots, L$, the element parameter  is mutated according to the following equation:
\begin{align*}
    \check{\theta}_{q}^{(u)}[h] = & {\rm bool}({\rm rand} \leq \check p_{q}) {\rm rand}  \\ &+ {\rm bool}({\rm rand} > \check p_q) \hat{\theta}_{q}^{(u)}[h]
\end{align*}
where $ \check p_q = \alpha_m (1- p_{i_q}  -p_{j_q}) $ is the mutation rate (
the constant $\alpha_m$ is the base mutation rate).  Therefore, 
the child parameter vectors with good parents with higher selection probabilities have lower mutation rates.  
  After the mutation process, the children population is generated as   $\check{\mathcal{G}}^{(u)} = \{\check{\theta}_1^{(u)},\cdots, \check{\theta}_L^{(u)}\} $.

\subsection{Evolution and stop}

Now, the next generation becomes
$\mathcal{G}^{(u+1)} =\check{\mathcal{G}}^{(u)}_1 \cup \mathcal{G}^{(u)}_2$
where the $L$ elements in $\check{\mathcal{G}}^{(u)}_1$ are from the offspring generation through crossover and mutation
and the $J-L$ elements in $\mathcal{G}^{(u)}_2$ are the leftover individuals.  
The evolutionary process  stops if there is no significant improvement of fitness between consecutive generations. Considering two consecutive generations $\mathcal{G}^{(u)}$ and $\mathcal{G}^{(u+1)}$, the stopping criterion is
\begin{align}
   |\Sigma_{i=1}^{J}    \delta_i^{(u+1)}- \Sigma_{i=1}^{J}  \delta_i^{(u)}|\leq \alpha_s
\end{align}
where $\alpha_s$ is a small positive constant called the evolution stop threshold.

\section{Experimental Evaluation}

Experimental results are reported in this section to compare the efficiency of
the proposed  GAAC algorithm with the conventional AC algorithm. 
The experiments were conducted on Mountain Car Continuous (MCC)-v0 and Swimmer-v3 \cite{purcell1977life}, two benchmarks from OpenAI Gym, which are elaborated in the following two subsections, respectively. 

\subsection{Mountain Car Continuous-v0}\label{expMountianCar}
The MCC environment consists of an underactuated car that starts its journey from the valley region between two hills. 
As to reach the flag present on top of the right hill, it must drive back and forth through the slope of the left and right hills to gain enough momentum to reach the goal. 
When the absolute value of the action that is applied to the car is larger, the reward is smaller (more negative). MCC is a sparsely rewarded environment where it only occasionally provides useful reward for the algorithm to leverage on. Here the reward remains always negative unless the car makes it to the flag. In that case, the car receives a +100 reward. 
More specifically, the state is $s_t = [x_t, y_t]^{\rm \scriptsize T}$ with $x_t$ being the car position and $y_t$ the speed,  
the action $a_t$ is the car acceleration (force), and $x_G=0.45$ the target position (i.e., position of the flag).  The reward function is defined as 
\begin{align*}
R(a_t, s_{t+1}) = \left\{ \begin{array}{ll}  -0.1 a_t^2, & x_{t+1} \neq x_G \\
+100, & x_{t+1} = x_G
\end{array} \right. .
\end{align*}
The action value is continuous within the range $[-1.0,1.0]$, out of which the value is clipped to its maximum or minimum value.  
For every episode, the initial position $x_1$ is set to a random value within the range $[-0.6,-0.4]$ and the initial 
speed $y_1=0$,
and the episode  runs and resets after running for $1,000$ steps. The episode may finish prematurely if the car reaches its goal sooner, i.e., once $x_{t+1} = x_G$ is achieved. 


The NN structures used in the experiments are same 
for all the algorithms and the parameters for AC and GAAC are summarized in Table~\ref{table:parameter}.

\begin{table}[h!]
\caption{Design parameters for AC and GAAC}
\centering
\begin{tabular}{ |p{8.5cm}|}
\hline
\multicolumn{1}{|c|}{\textbf{AC Algorithm}}\\
\hline
\textbf{Actor NN:} 
2 hidden layers;  $40/400^*$ neurons each layer; learning rate $\alpha_a = 0.00001/0.001$ \\
\hline
\textbf{Critic NN:} 2 hidden layers; 400 neurons each layer; learning rate $\alpha _c = 0.00056/0.0001$;
discount factor $\gamma= 0.99$ \\
\hline
\multicolumn{1}{|c|}{\textbf{GAAC Algorithm}}\\
\hline
\textbf{Stage 1:} \# of round $M= 10$; \# of episodes per round $N = 3/6$; \# of total episodes $E=30/60$\\
\hline
\textbf{Stage 2 (PFM):}  2/3 hidden layers; 40/64 neurons each;  ELU activation;  Xavier / Glorot normal weight initialization \\
\hline
\textbf{Stage 2  (GA):} $\eta = 25\%/15\%$;  population size $J=50$; \# of parents $K= 25$;  \# of generations 20;
base mutation rate $\alpha_m=0.01$;
evolution stop threshold  $\alpha_s = 0.1/0.01$\\
\hline
\textbf{Stage 3:}   optimal policy NN: 2/4 hidden layers; 40/400 neurons each layer\\
\hline
\end{tabular}
\label{table:parameter}
\begin{flushleft}
\vspace{-2mm}
{  $^*$ The two values of a parameter are for MCC and Swimmer respectively, i.e., 
MCC/Swimmer.  The one-valued parameters are common for both environments. }
\end{flushleft}
\end{table}

\begin{figure}[t]
    \centering
    \begin{subfigure}{\columnwidth}
        \centering
    \includegraphics[width=0.8\columnwidth]{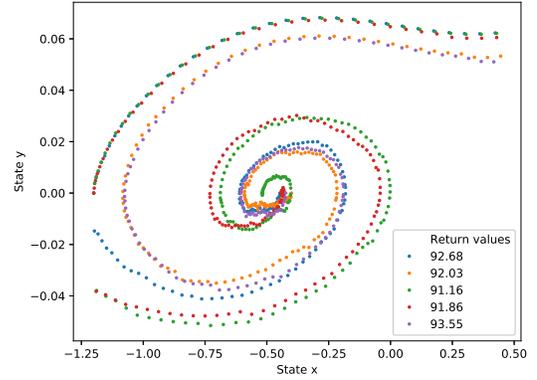}
    \caption{}
    \label{fig.AC_a}
\end{subfigure}
    \begin{subfigure}{\columnwidth}
        \centering
        \includegraphics[width=0.8\columnwidth]{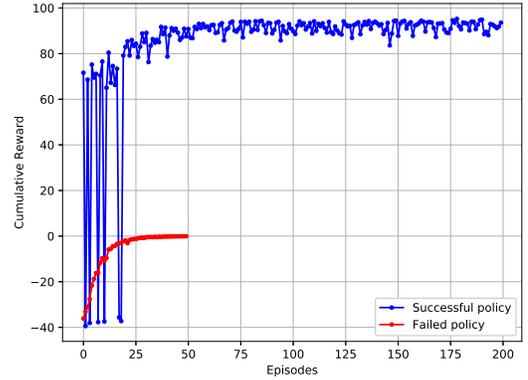}
        \caption{}
        \label{fig.AC_b}
\end{subfigure}
    \caption{(a) Illustration of the car's trajectories (each dot represents
    the car's instantaneous position $x$ and speed $y$ and the dots in the same color make one episode)
     for the final 5 episodes of the 200 episodes demonstrating successful policy exploration using AC. 
They all reach the target position but with different final speeds. The cumulative reward obtained
in each episode is recorded in the legend.  (b) 
Illustration of cumulative reward vs episode
(each dot represents the cumulative reward of one episode.) Failed episodes appeared in the early stage and then successful ones dominated, which verifies the effectiveness of the AC algorithm.
}
    \label{fig.AC}
\end{figure}

\begin{figure}[t]
\centering
\begin{subfigure}{\columnwidth}
\centering
\includegraphics[width=0.8\columnwidth]{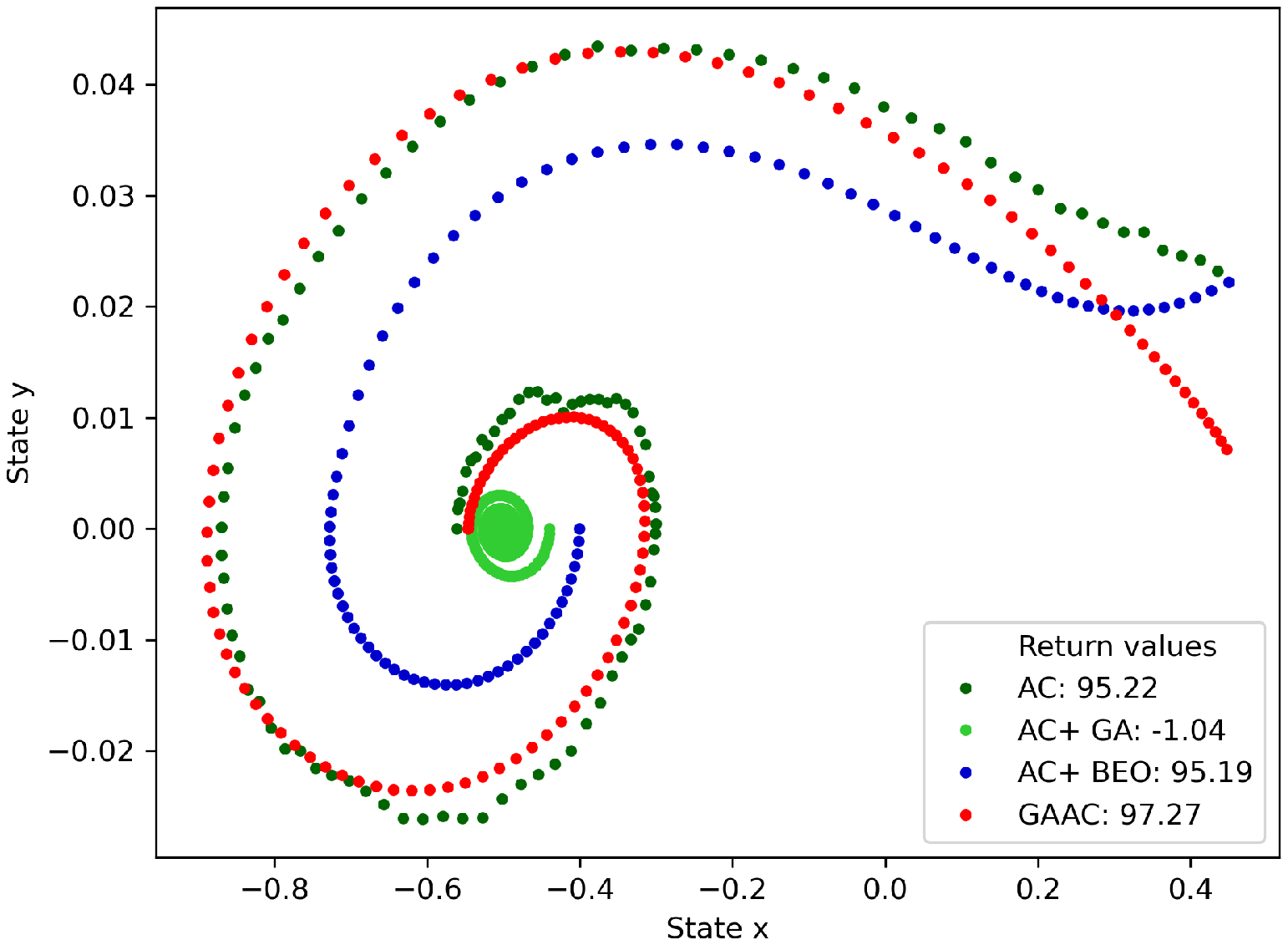}
\caption{}
\label{fig:comp_a}
\end{subfigure}
\begin{subfigure}{\columnwidth}
\centering
\includegraphics[width=0.8\columnwidth]{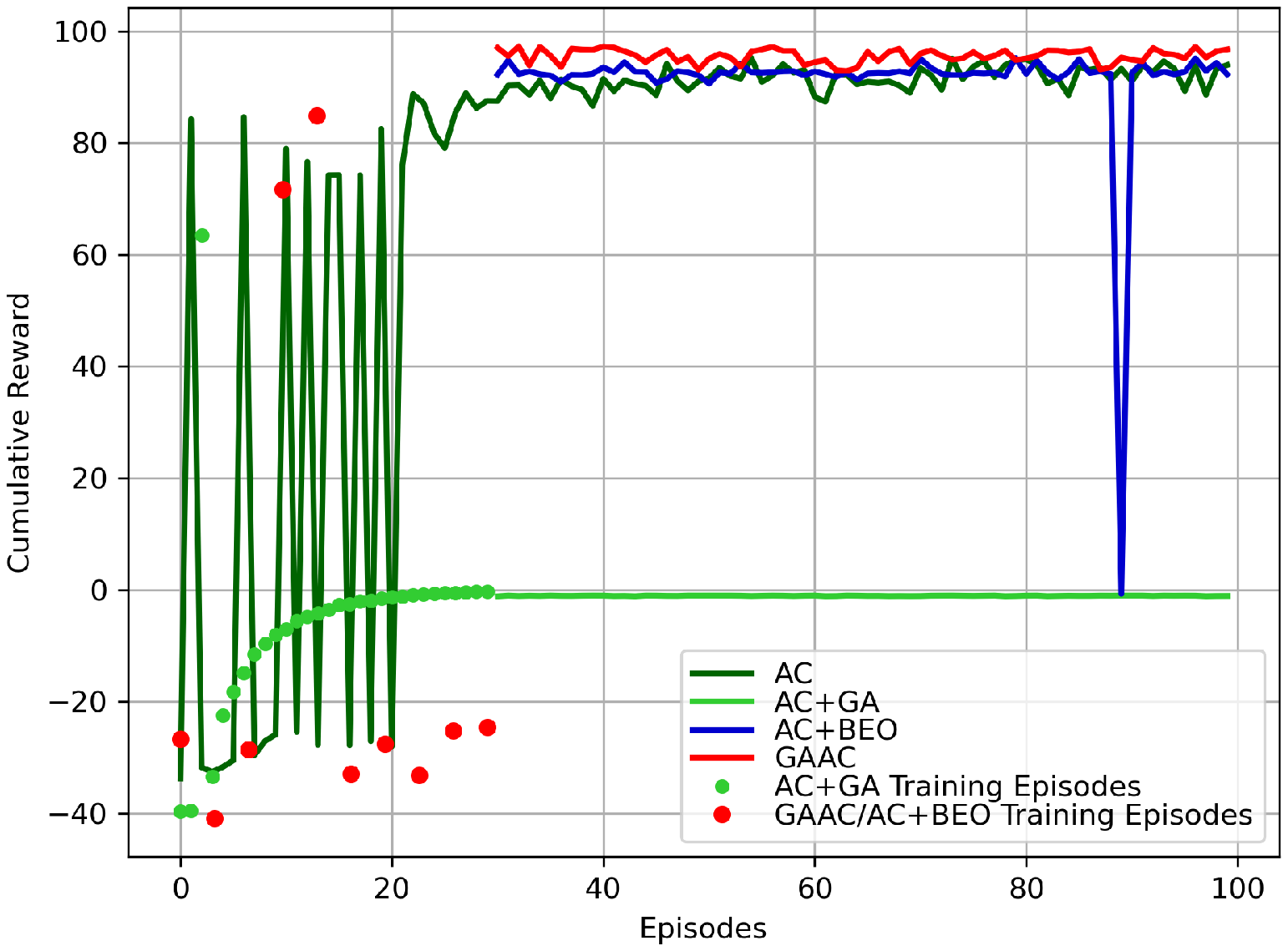}
\caption{}
\label{fig:comp_b}
\end{subfigure}
\caption{
(a) Illustration of the car's trajectories  
achieved by the four algorithms with the cumulative rewards  recorded in the legend.
The  AC+ GA, AC+BEO and GAAC episodes are the best testing ones after learning from  30 episodes
and the AC episode is the one learnt online for over 100 episodes.  (b) Comparative illustration of
cumulative reward vs episode for the four algorithms. }
\label{fig:comp}
\end{figure}

\subsubsection{Successful and failed episodes using AC}

The AC algorithm starts with a policy of random initial parameters. 
It may reach the target and win a reward of +100  in an episode (called a successful episode) or get ``stuck at
local minima'' (called a failed episode), and sometimes the policy cannot recover from such episodes leading to a failed policy (marked in red), see Fig.~\ref{fig.AC_b}. 
If the AC algorithm attains a successful episode in the first few trials, then with every episode the policy is expected to gradually improve the cumulative result with increasing rewards. 
It was observed that in about  200 episodes a cumulative reward average of 92.8
was obtained and it was increased to 94.2 in 5,000 episodes.  Fig.~\ref{fig.AC_a} shows the trajectories of the car for the 200 episode trained AC policy. The car was able to  reach the goal position at $x_G=0.45$ but the final 
speeds were relatively large.
The performance of the AC is plotted in Fig.~\ref{fig.AC_b} for the first 200 episodes. 
The cumulative reward of an episode falls short of $95$.
Both successful and failed episodes can be observed in the figure.

\subsubsection{Optimization of training datasets}

The critical mechanism of the proposed GAAC approach is 
optimization of training datasets.  
We first conducted the AC algorithm for $M=10$ rounds with $N=3$ episodes each. 
And we chose the best episode of the round which gives the highest cumulative reward. 
At the onset of each round, we reset the networks and run the AC policy from scratch.  
 Out of the ten best episodes, there was one good episode
 in which the target was achieved and the other nine were bad. 
In the  ten episodes,  we collected totally  $|{\cal D}_1| =9,654$ samples where the operator 
$|\cdot|$ represents the cardinality of a set. 
It is noted that a failed episode typically has more samples than a good one 
as the latter may stop earlier once the target is reached.  In average, 
each episode contributes $965$ samples.
Next, the GA module optimized $\eta =25\%$ of the total samples, i.e., $|{\cal D}'| = 2,416$. 
Then, the optimized dataset was used for training the optimal policy network. 
Another 80 episodes were tested on the trained policy and only one episode failed. 
The  cumulative rewards were located in the range between $92$ and $93$ for good episodes. 
The results are summarized  in Table~\ref{table:samples}. 
The same process was repeated and out of the ten best episodes there were two good episodes and
eight bad ones.  Their results are also recorded in the same table for comparison. With more good samples, all of
the 80 testing episodes were successful and the cumulative rewards were improved to 
the range between $95$ and $98$. 
 It is worth mentioning that the successful/failed episode ratio is 1:9 or 2:8 in the experiments 
 summarized in Table~\ref{table:samples} because successful policies are rare in MCC in the early stage of 
 the AC algorithm for the local minima issue. The case with 2:8 is used in the subsequent discussion.

\begin{table}[h!]
\centering
\caption{Evaluation of optimal policy under different successful  and failed training episode mixture ratios in MCC}
\begin{tabular}{ |p{1.3cm}|p{1.3cm}|p{1.3cm}|p{1.3cm}|p{1.3cm}| }
\hline
 episode mixture ratio & $|{\cal D}_1|$: samples by BEO & $|{\cal D}'|$:  GA optimized  samples&  
failed testing episodes & Cumulative Reward range\\
\hline
1:9 & 9,654 & 2,416 &   1 out of 80 & 92 - 93 \\
\hline
2:8 & 8,894   & 2,224   & 0 out of 80 & 95 - 98 \\
\hline
\end{tabular}
\label{table:samples}
\end{table}

\subsubsection{ Ablation studies} \label{MCC_GAAC artifacts}
The effectiveness of the design is evaluated through ablation studies of four algorithms. 
The first one is the conventional {\bf AC}
algorithm where the dataset collected from the actor NN is $\mathcal{W}_o$ 
from the which the  optimal policy is directly trained. 
The second one is a partial algorithm of GAAC, called {\bf AC+GA}, where the AC explored data $\mathcal{W}_o$ is directly used for 
GA optimization and training of the optimal policy network in stage 2 and stage 3, respectively, that is, 
the BEO component in stage 1 is excluded.
The third one is another partial algorithm  including stage 1 and stage 3, but not stage 2, 
called {\bf AC+BEO}.  In other words,  the dataset $\mathcal{W}_1$ from stage 1 is directly used in stage 3 for training 
the optimal policy network. 
The fourth one is the full three-stage AC algorithm with both BEO and GA module, 
 i.e., {\bf GAAC}, where  the dataset $\mathcal{W}_2$  is used for training the optimal policy network.

Performance comparison among the four algorithms is demonstrated in Fig.~\ref{fig:comp}.The result from the AC algorithm has been explained in  Fig.~\ref{fig.AC}.  In the AC+GA algorithm, due to the absence of the BEO module, all the $30$ training episodes were from a single continuously learning AC policy (i.e., one round).  As mentioned before, successful policies are rare in MCC in the early stage of  the AC algorithm, so a failed policy is more likely and recorded here. 
The GA refined samples from these training episodes were not effective for learning an optimal policy.
The AC+ BEO algorithm learnt the policy using the samples collected from only  $30$ episodes
even though the cumulative reward in the range between $92$ and $93$ did not significantly outperform the AC algorithm. 
It is worth mentioning that these 30 episodes were from ten rounds in the BEO stage.  The BEO mechanism for using a small percentage (two out of ten) of successful polices
demonstrated its effectiveness in resolving the aforementioned local minima issue.

Finally, the complete GAAC algorithm was  implemented using the same 30 episodes as in AC+BEO.
In Fig.~\ref{fig:comp_a}, the GAAC trajectory shows that the car was able to 
reach the goal position with a lower speed.
The result in Fig.~\ref{fig:comp_b} shows that the GAAC algorithm performs better than AC, AC+GA and AC+BEO.
It achieved an average cumulative reward of $95.83$ over $80$ testing episodes, again using the samples collected from only  30 episodes. 
For the conventional AC algorithm, it took  more than 5,000 episodes to attain the same level of optimality in terms 
of the average cumulative reward.

\begin{figure}[t]
 \centering
\includegraphics[width=0.8\columnwidth]{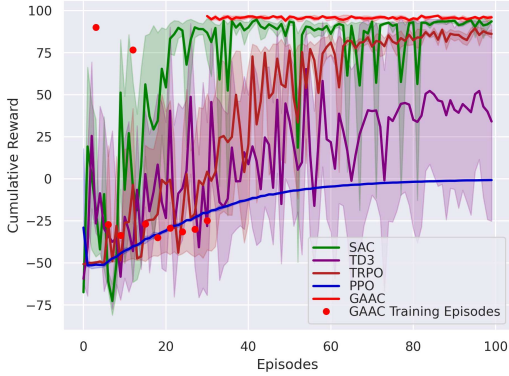}
 \caption{Performance comparison of the GAAC algorithm with the existing benchmarks in MCC experiments.}
 \label{fig.bench_compare1}
\end{figure}

\subsubsection{Comparison with other benchmarks} To further evaluate the performance of GAAC, we tested it against some of the latest benchmarks, 
e.g.,  SAC \cite{haarnoja2018soft}, TD3 \cite{fujimoto2018addressing}, TRPO \cite{schulman2015trust} and PPO\cite{schulman2017proximal}.
We used a baseline library called stable-baselines \cite{stable-baselines} for generating the data for the benchmark. 
For each algorithm, we used the data from five repeated tests of the policy. 
 The experimental results from the algorithms are plotted in Fig.~\ref{fig.bench_compare1} that shows a solid mean line surrounded by a lightly shaded area representing its variance in the five repeated tests. The training episodes for the five repeated tests are represented by one dot as the average reward for clarity.
For the GAAC algorithm, only the mean of the training data is shown for the first 30 episodes for neat presentation. The benchmarks use the default hyperparameters of the stable-baselines library. They also use a multilayered policy network similar to GAAC. 

{ 
All of the benchmark algorithms frequently suffer from the local minima issue in MCC as discussed before. 
For the purpose of comparison, we only selected the successful policies in Fig.~\ref{fig.bench_compare1}
except the PPO algorithm that failed to learn a successful policy.} 
The plots show that  GAAC attained a higher level of optimality and faster convergence to the optima by consuming the data from only 30 episodes.  So, GAAC learned an optimal policy with significantly less data samples from the environment than the existing benchmarks in this comparison. 

The $31$st-$100$th episodes in Fig.~\ref{fig.bench_compare1} are called the evaluation episodes. 
So, there are 350 evaluation episodes recorded in figure from the five repeated tests. 
The quantitative comparison is also summarized in Table~\ref{table:quant_comp1} in terms of the mean and standard variation 
of the rewards for these $350$ evaluation episodes. It concludes that GAAC outperforms the benchmarks
by achieving the highest reward of $95.684\pm 1.146$.

\begin{table}[h!]\centering
\caption{Rewards of the evaluation episodes}
 \begin{tabular}{|l|l|l|}
\hline\hline
\textbf{Algorithm} & MCC-v0 & Swimmer-v3\\
\hline \hline
\textbf{SAC} & $86.243\pm 27.759$ &  $ 21.879 \pm 16.333 $ \\
\textbf{TD3} & $26.245 \pm 54.071$  &   $ 24.495 \pm 23.065$ \\
\textbf{TRPO} & $66.486 \pm 38.092$  & $ 26.380 \pm 8.961$  \\
\textbf{PPO} & $-5.892\pm 5.309$  &  $32.182 \pm 3.631$  \\
\textbf{AC} & $ 90.440 \pm 25.434 $ &  $7.189 \pm13.887$  \\
\textbf{GAAC} & $\mathbf{95.684\pm 1.146} $ &  $\mathbf{78.560 \pm 29.914}$  \\
\hline \hline
\end{tabular}
\label{table:quant_comp1}
\end{table}


\subsection{Swimmer-v3}\label{expSwimmer}
Swimmer-v3 represents a planar robot swimming in a viscous fluid. It is made up of three links 
(head, body and tail) and two actuated joints connecting them. 
The system dynamics can be described in a ten-dimensional state space, which consists of 
position and velocity of the center of the body (4), the angle and angular velocity of center of body (2), and
angle and angular velocity of the two joints (4). 
The two-dimensional action space consists of the torques applied on the two actuated joints. 


The objective in this experiment is to stimulate the maximal forward (the positive x-axis) moving/swimming by actuating the two joints,
with the reward function defined as follows
\begin{align*}
    R(a_t,s_{t+1}) &= v^{x}_{t+1} - 0.0001\, \|a_{t}\|^2_2 
\end{align*}    
where $v^{x}$ (an element of $s$) is the forward velocity  and $a_{t}$ the two-dimensional action torques.  
The value of an action torque is continuous within the range $[-1.0,1.0]$, out of which the value is clipped to its maximum or minimum value.  
For every $1,000$ steps, called an episode, the environment is reset and the swimmer starts at a new random initial state. 
There is no premature termination condition applied to an episode.

\begin{figure}[t]
\centering
    \begin{subfigure}{\columnwidth}
            \centering
	\includegraphics[width=0.8\columnwidth]{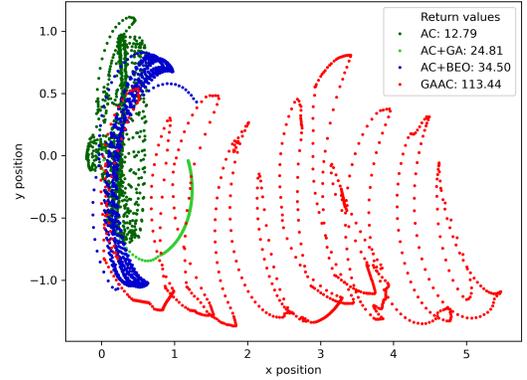}
	        \caption{}
	        \label{subfig:pos_plot}
\end{subfigure}
    \begin{subfigure}{\columnwidth}
            \centering
	\includegraphics[width=0.8\columnwidth]{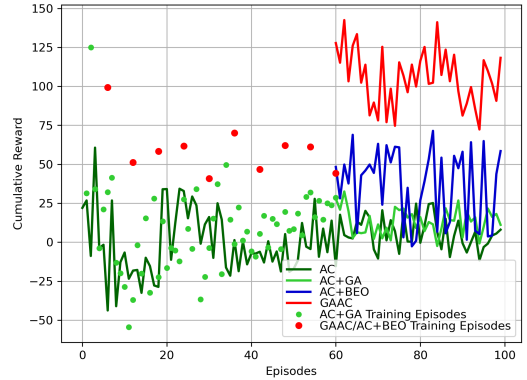}
	        \caption{}
		\label{subfig:comp_plot}
	\end{subfigure}
\caption{
(a) Illustration of the swimmer motion trajectories  
achieved by the four algorithms with the cumulative rewards  recorded in the legend.
The  AC+ GA, AC+BEO and GAAC episodes are the best testing ones after learning from 60 episodes
and the AC episode is the one learnt online for over 100 episodes.  (b) Comparative illustration of
cumulative reward vs episode for the four algorithms. }
\label{fig:basic_compare}
\end{figure}

\subsubsection{Optimization of training dataset}\label{swimerOptTrainData}
 We conducted the AC policy for $M = 10$ rounds with $N=6$ episodes per round. Choosing the best episode in each round, we collected $|\mathcal{D}_{1}| = 10,000$ samples. Each episode here contributes equally $1,000$ samples. Next the GA module optimizes $\eta=15\%$ of total samples, i.e. $|\mathcal{D}'| = 1,500$. The design parameters for AC and GAAC in the Swimmer environment are also summarized in Table \ref{table:parameter}.

\subsubsection{ Ablation studies}  \label{Swim_GAAC_artifacts}
In Fig.~\ref{subfig:pos_plot}, we compare the motion trajectories of the swimmer's center of body in the x-y plane, using the four mentioned algorithms, i.e., AC, AC+BEO, AC+GA and GAAC. It is observed that the swimmer performed better in maximizing its  forward velocity along the x-axis, using the actions sampled from the GAAC trained policy. 
In particular, the swimmer with the GAAC policy was able to achieve a reasonable forward locomotion 
in 1,000 steps of one episode, while  that with other policies was not.  
In Fig.~\ref{subfig:comp_plot}, we compare the four algorithms for their performance measured in terms of the cumulative rewards attained per episode.  It is evident from the figure that AC+GA  performed only marginally better than the conventional AC algorithm, but AC+BEO improved  AC by increasing the mean performance from around 15 to 30.  Furthermore, GAAC improved the performance to be above 75.

\subsubsection{ Percentage of samples for GA optimization} \label{swimerOptbeta}
\begin{figure}[t]
\centering
\includegraphics[width=0.8\columnwidth]{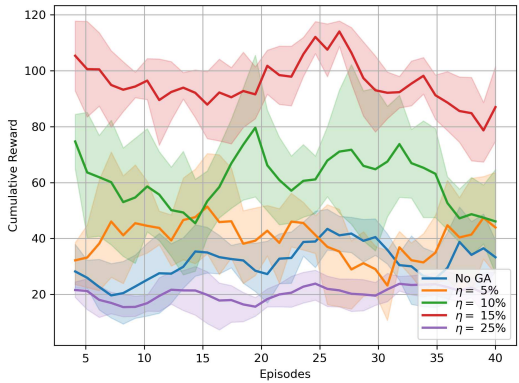}
\caption{ Performance of the GAAC algorithm with different 
percentage ($\eta$) of samples for GA optimization in Swimmer-v3.}
\label{fig: GA upd compare}
\end{figure}

For the GAAC algorithm used in Fig.~\ref{fig:basic_compare}, the percentage of samples
for GA optimization was set as $\eta=15\%$. 
More experiments were done in order to understand the influence of this hyperparameter. 
Using the same training episodes, the GA algorithm was implemented with different $\eta$, repeated for five times, and the results were recorded in Fig~\ref{fig: GA upd compare}.  
The figure shows a solid mean line surrounded by a lightly shaded area representing its variance in the five repeated tests
 and the training episodes ($1$st - $60$th) are not plotted for clarity. 
 The results clearly show the influence of $\eta$ and the best choice is $\eta=15\%$ for Swimmer-v3.
They also indicate that improving too few samples with GA is insufficient for effective training of the policy network, but improving too many samples may also cause loss of diversity resulting in poor performance.
        
\subsubsection{Comparison with other benchmarks}
\begin{figure}[t]
\centering
\includegraphics[width=0.8\columnwidth]{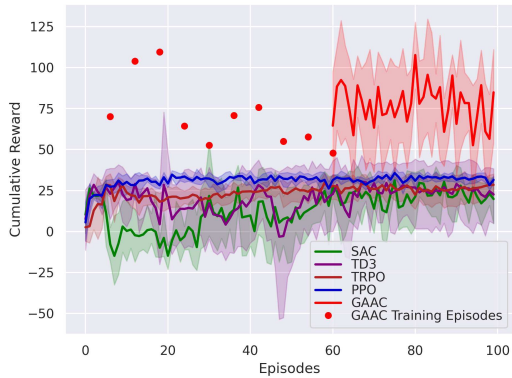}
\caption{Performance comparison of the GAAC algorithm with the existing benchmarks in Swimmer experiments.}
\label{fig: bench compare}
\end{figure}

The comparison of GAAC with other benchmark algorithms is presented in Fig.~\ref{fig: bench compare}. The plots show that the GAAC algorithm produced substantially better performing policies in this high dimension environment,  with very few (60) learning episodes. Indeed, GAAC, trained with 60 episodes, outperformed the benchmark algorithms even after they have been trained for more than 500 episodes, i.e., 50,000 steps (not shown in the figure). 
The outstanding performance of GAAC is also demonstrated by 
the quantitative comparison summarized in Table~\ref{table:quant_comp1} for 
the 200 evaluation episodes, i.e., the $61$st-$100$th episodes repeated five times.

\section{Conclusion and Future Work}

In this paper, we have proposed an optimal AC policy with a GA optimized training dataset.
The process is made of a best episode only operation, a policy parameter-fitness model,
and a genetic algorithm module.   
The new approach can learn an optimal policy with significantly less number of samples compared to the latest benchmarks, thus  demonstrating the improvement in 
sampling efficiency and convergence speed over the conventional  AC algorithm.  In this work, GAAC is evaluated in two dynamic control environments with different state and action dimensions and  its superiority is exhibited. It is an interesting future work to apply the proposed algorithm 
in different types of control tasks in more challenging environments. Moreover, the idea of optimizing training dataset can be integrated with other advanced RL algorithms like SAC, TD3, etc. Improvement with hyper-parameter tuning techniques for neural networks and 
deep neural networks will be other interesting topics for future research. 
%
%
\bibliographystyle{IEEEtran}
\bibliography{refer}
\end{document}